\RequirePackage{amsmath}
\documentclass[runningheads]{llncs}
\usepackage[T1]{fontenc}
\usepackage{graphicx}
\usepackage{amssymb}
\usepackage{bm}
\usepackage{booktabs}
\usepackage{tabularx}
\usepackage[table]{xcolor}

\DeclareMathOperator{\expected}{\mathbb{E}}
\DeclareMathOperator{\probability}{\mathbb{P}}

\newcommand{\figref}[1]{\figurename~\ref{#1}}
\newcommand{\tabref}[1]{\tablename~\ref{#1}}

\usepackage{hyperref}
\hypersetup{
    colorlinks,
    linkcolor={blue!50!black},
    citecolor={blue!50!black},
    urlcolor={blue!50!black}
}

\begin{document}

\title{Average Calibration Error: A Differentiable Loss for Improved Reliability in Image Segmentation}
\titlerunning{Average Calibration Error: A Differentiable Loss for Reliable Segmentation}

\author{Theodore Barfoot\inst{1} \and  
	Luis C. Garcia Peraza Herrera\inst{1} \and  
	Ben Glocker\inst{2}  \and \newline 
	Tom Vercauteren\inst{1}} 


\authorrunning{T. Barfoot et al.}

\institute{King's College London, London, UK \\
	\email{theodore.d.barfoot@kcl.ac.uk}, \email{luis\_c.garcia\_peraza\_herrera@kcl.ac.uk}, \email{tom.vercauteren@kcl.ac.uk} \and
	Imperial College London, London, UK \\
	\email{b.glocker@imperial.ac.uk}}
\maketitle
%
%
%
\begin{abstract}
Deep neural networks for medical image segmentation often produce overconfident results misaligned with empirical observations.
Such miscalibration challenges their clinical translation.
We propose to use marginal L1 average calibration error (mL1-ACE) as a novel auxiliary loss function to improve pixel-wise calibration without compromising segmentation quality.
We show that this loss, despite using hard binning, is directly differentiable, bypassing the need for approximate but differentiable surrogate or soft binning approaches.
Our work also introduces the concept of \emph{dataset reliability histograms} which generalises standard reliability diagrams for refined visual assessment of calibration in semantic segmentation aggregated at the dataset level.
Using mL1-ACE, we reduce average and maximum calibration error by 45\% and 55\% respectively, maintaining a Dice score of 87\% on the BraTS 2021 dataset.
We share our code here: \href{https://github.com/cai4cai/ACE-DLIRIS}{https://github.com/cai4cai/ACE-DLIRIS}.
\end{abstract}
\section{Introduction}

Deep neural networks (DNNs) have significantly advanced the field of semantic segmentation.
However, their large capacity makes them susceptible to overfitting, leading to overconfident predictions that do not accurately reflect the underlying uncertainties inherent to the task \cite{Guo2017}.
In medical image segmentation, where the confidence of predictions can be as crucial as the predictions themselves, and overfitting is exacerbated by smaller datasets, such overconfidence poses significant risks, particularly when this miscalibration is not communicated to the end-user.
Appropriate management of confidence
can also have benefits for patient outcomes, for example in radiotherapy contouring \cite{Bernstein_2021}.

The Dice Similarity Coefficient (DSC) loss is a popular choice in medical image segmentation due to its robustness against class imbalance \cite{lossodyssey2021}, but it is known to yield poorly calibrated, overly confident predictions \cite{Mehrtash2020confidence}.  

\begin{figure}[tbp]
	\centering
	\def\svgwidth{1\textwidth}
\begingroup%
  \makeatletter%
  \providecommand\color[2][]{%
    \errmessage{(Inkscape) Color is used for the text in Inkscape, but the package 'color.sty' is not loaded}%
    \renewcommand\color[2][]{}%
  }%
  \providecommand\transparent[1]{%
    \errmessage{(Inkscape) Transparency is used (non-zero) for the text in Inkscape, but the package 'transparent.sty' is not loaded}%
    \renewcommand\transparent[1]{}%
  }%
  \providecommand\rotatebox[2]{#2}%
  \newcommand*\fsize{\dimexpr\f@size pt\relax}%
  \newcommand*\lineheight[1]{\fontsize{\fsize}{#1\fsize}\selectfont}%
  \ifx\svgwidth\undefined%
    \setlength{\unitlength}{345.82676625bp}%
    \ifx\svgscale\undefined%
      \relax%
    \else%
      \setlength{\unitlength}{\unitlength * \real{\svgscale}}%
    \fi%
  \else%
    \setlength{\unitlength}{\svgwidth}%
  \fi%
  \global\let\svgwidth\undefined%
  \global\let\svgscale\undefined%
  \makeatother%
  \begin{picture}(1,0.47540986)%
    \lineheight{1}%
    \setlength\tabcolsep{0pt}%
    \put(0,0){\includegraphics[width=\unitlength]{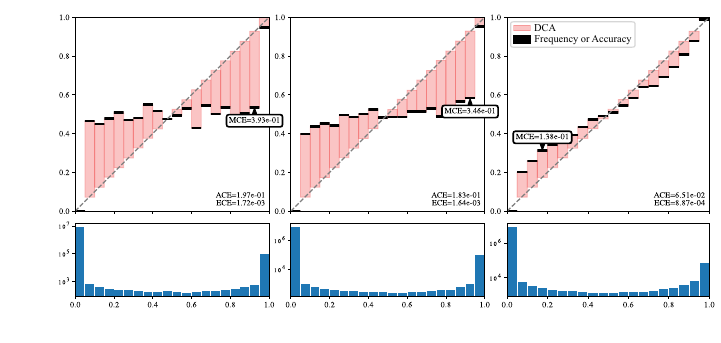}}%
    \put(0.25,0.012){\color[rgb]{0,0,0}\makebox(0,0)[lt]{\lineheight{1.25}\smash{\begin{tabular}[t]{l}Predicted Foreground Probability / Confidence\end{tabular}}}}%
    \put(0.065,0.07){\color[rgb]{0,0,0}\rotatebox{90}{\makebox(0,0)[lt]{\lineheight{1.25}\smash{\begin{tabular}[t]{l}Count\end{tabular}}}}}%
    \put(0.035,0.19){\color[rgb]{0,0,0}\rotatebox{90}{\makebox(0,0)[lt]{\lineheight{1.25}\smash{\begin{tabular}[t]{l}Empirical Foreground\end{tabular}}}}}%
    \put(0.065,0.19){\color[rgb]{0,0,0}\rotatebox{90}{\makebox(0,0)[lt]{\lineheight{1.25}\smash{\begin{tabular}[t]{l}Frequency / Accuracy\end{tabular}}}}}%
  \end{picture}%
\endgroup%
	\caption{Image-specific reliability diagrams for whole tumour segmentation (BraTS case 00054) under different training losses.
	Left: DSC; Middle: DSC + temperature scaling (Ts); Right: DSC + mL1-ACE (proposed).
	\label{fig:case_reliability_diagrams_dice}}
\end{figure}
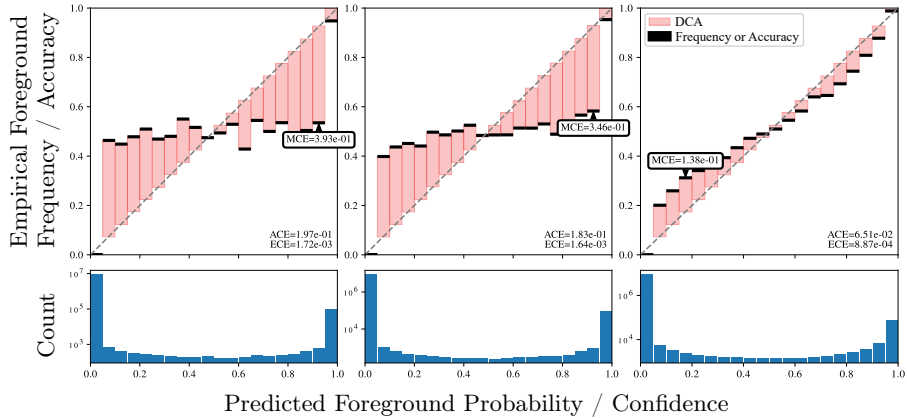

Reliability diagrams are the reference means of assessing and visualising the calibration performance of a predictor over a dataset and this has extensively been used for classification tasks~\cite{DeGroot1983}.
They are constructed by binning the samples according to a discretisation of the predicted probabilities, also referred to as confidence. Within each bin, the average confidence and average accuracy is computed.
The reliability diagram is obtained by plotting the Difference between Confidence and Accuracy (DCA) as a function of confidence.
An example is shown in \figref{fig:case_reliability_diagrams_dice}.
An aggregation of these DCA values can then be used to estimate measures of calibration error.
While reliability diagrams and associated calibration error metrics are typically computed at the dataset level, 
in semantic segmentation, these diagrams can also be constructed per image, due to multiple pixels/voxels being present.
Capturing both image-specific and dataset-wide reliability is of particular relevance in medical imaging, due to the safety critical nature of the task. 
Box-plot diagrams have been used in combination with image-specific reliability diagrams to represent calibration variability within an image segmentation dataset \cite{kock2022histograms}.

Traditional post-hoc calibration strategies such as temperature scaling (Ts) \cite{Guo2017,platt1999probabilistic} offer a partial remedy by adjusting confidence levels uniformly.
While more nuanced post-hoc calibration methods offer class-specific adjustments~\cite{kull2019dirichlet,ding2020local,islam2021class},
they still fail to exploit the extensive parameter space of DNNs~\cite{liang2020imporved,hebbalaguppe2022stitch}.

Train-time calibration methods
are more effective in improving calibration than post-hoc methods \cite{Guo2017}.
They leverage the large model capacity in DNNs to provide fine-grained calibration adjustment with no detriments for the primary task.
For image classification tasks, DCA-based auxiliary losses such as expected calibration error (ECE) have been shown to improve calibration \cite{liang2020imporved}.
However, the accuracy term in the DCA-based losses is not differentiable due to the argmax operator.
Furthermore, the usage of hard binning in standard DCA has led to previous work seeking for alternative differentiable calibration losses \cite{liang2020imporved}.
Multi-class DCA (MDCA) avoids binning, and its associated differentiability challenges, by calculating DCA averaged over each class within a mini-batch rather than averaged within confidence bins \cite{hebbalaguppe2022stitch}.
A differentiable surrogate for expected calibration error (DECE), has been proposed \cite{bohdal2023metacalibration}.
However, the DECE approach relies on an approximation of accuracy, as well as soft binning to avoid the non-differentiability of hard binning.
The DECE work also exploits non-trivial meta-learning rather than direct optimisation of the loss
thereby complicating its use and limiting its wider applicability.

In a different stream of work, authors have
introduced specialised loss functions which do not rely on DCA methods to tackle overconfidence in biomedical image segmentation.
For example, DSC\texttt{++} loss, which uses a modified DSC loss to enhance calibration by adjusting the false positive (FP) and false negative (FN) terms of DSC based on class frequency \cite{yeung2021calibrating}.
Neighbor-Aware Calibration (NACL) is another example focusing on spatial consistency, via penalty constraints on logits, to improve reliability and performance in segmentation tasks~\cite{murugesan2023neighbor,murugesan2024neighbor}.
While these methods have demonstrated some improvement in calibration, their indirect approach make them more difficult to interpret and potentially lead to suboptimal results. 

In this work, we present a novel auxiliary loss function, marginal L1 Average Calibration Error (mL1-ACE), designed to enhance calibration accuracy in medical image segmentation without compromising on segmentation quality.
This represents the first clear attempt to use established calibration metrics directly as an auxiliary loss in the context of medical image segmentation.
Despite presumptions related in previous DCA-based methods, our approach is inherently differentiable, even with hard-binning of probabilities, eliminating the need for surrogates or soft-binning techniques.
mL1-ACE employs a class-wise calibration strategy, opting for ACE over ECE to better address the class imbalances in medical datasets.

While the concept of average calibration error (ACE) is not new, our key contribution lies in adapting it as a directly differentiable auxiliary loss for semantic segmentation, a novel approach not addressed in prior literature. This provides a major distinction from the use of ACE as a metric in classification tasks. The transition from classification to segmentation is a key aspect which we are the first to identify, enabling ACE to be used as a loss rather than a metric. Classification training pipelines provide too few samples within a mini-batch to meaningfully compute its ACE, in contrast to segmentation where each voxel provides a sample.

When combined with standard loss functions, in particular DSC, mL1-ACE demonstrates superior calibration performance on the BraTS2021 dataset \cite{BraTS2021}, significantly surpassing conventional post-hoc calibration methods like Ts.
Additionally, we introduce dataset reliability histograms, offering a comprehensive visual tool for evaluating and communicating model calibration across image segmentation datasets, enhancing transparency in reporting model reliability beyond predictive performance.
\section{Method}
For calculation of image-specific calibration metrics, and our subsequent mL1-ACE loss, we discretise the continuous predicted probabilities of each of the $C$ classes into $M$ bins.
For any given image, and class $c$, the expected foreground probability $e^c_m$ within bin $m$ (also referred to as confidence in DCA) is compared to the observed frequency $o^c_m$ within that same bin (also referred to as accuracy in DCA).
This provides us with an analogous representation of DCA but extended to multi-class setting.
An ideally calibrated model is one where the gaps, $|o^c_m - e^c_m|$ equal zero over all $M$ bins and all $C$ classes.
That is, a perfectly calibrated model is one where the predicted foreground probabilities align precisely with the observed frequency of that class across numerous samples.
For example, if a well-calibrated brain tumour segmentation model segments $100$ voxels, each with a probability/confidence of $0.7$, then we expect $70$ of those voxels to be tumour.

Expected Calibration Error (ECE) \cite{Naeini2015} distils the reliability diagram described above into a single scalar value that can be used to assess the calibration of the model, by considering the weighted sum of the absolute difference of the gaps over all the bins.
We extend this approach to a multi-class and image-specific variant, which we refer to as marginal L1 expected calibration error:
\begin{equation}
	\textrm{mL1-ECE} = \frac{1}{C} \sum_{c}^{C} \sum_{m}^{M} \frac{n^c_m}{N} \lvert o^c_m - e^c_m \rvert
	\label{eq:ece}
\end{equation}
where
$n^c_m$ is the number of in bin $B^c_m$, the $m^{\textrm{th}}$ bin for class $c$,
and $N$ is the total number of voxels for an image.
More formally, we define the observed foreground frequency $o^c_m$ and the expected foreground probability $e^c_m$ as:
\begin{equation}
	o^c_m = \probability[\bm{Y}^c_i \,\rvert\, i \in B^c_m], \quad e^c_m = \expected[f_{\bm{\theta}}(\bm{X})^c_i \,\rvert\, i \in B^c_m]
	\label{eq:o}
\end{equation}
where $\bm{X}$ is the input image,
$\bm{Y}$ the associated ground truth label-map,
and $f_{\bm{\theta}}$ is our predictor with trainable weights $\theta$.

Average Calibration Error (ACE) \cite{neumann2018a}, performs this distillation in a similar way, but considers an unweighted sum over all bins.
We extend this approach to a multi-class and image-specific variant, which we refer to as marginal L1 average calibration error:
\begin{equation}
	\textrm{mL1-ACE} = \frac{1}{CM} \sum_{c}^{C} \sum_{m}^{M}  \lvert o^c_m - e^c_m \rvert
	\label{eq:ace}
\end{equation}

Finally, we report Maximum Calibration Error (MCE) \cite{Naeini2015} which is only concerned with the maximum gap but extend it to a multi-class and image-specific variant while referring to it as marginal L1 maximum calibration error:
\begin{equation}
	\textrm{mL1-MCE} = \frac{1}{C} \sum_{c}^{C} \max_{m}  \lvert o^c_m - e^c_m \rvert
	\label{eq:mce}
\end{equation}

We anticipate that 
all three measures, as shown in equations \eqref{eq:ece}, \eqref{eq:ace}, and \eqref{eq:mce}, should be differentiable almost everywhere.
Combined with their image-specific nature, this characteristic
makes them trivially usable as auxiliary losses that can be paired with any image segmentation loss.
Specifically, useful differentiability arises due to the large number of voxels that have membership to each bin with averaging being performed within each bin rather than across the bins themselves.
However, given that mL1-MCE only considers the maximum gap, for a single bin, it is less likely to provide a sufficient supervisory signal compared to mL1-ECE and mL1-ACE.

Since mL1-ECE is weighted by the relative proportion of voxels in each bin, if a model is performing sufficiently well, mL1-ECE is heavily skewed towards the first and last bin as is evident in \figref{fig:case_reliability_diagrams_dice}.
Therefore, we argue that mL1-ACE is the most appropriate measure for calibration, and therefore an excellent candidate for an auxiliary loss targeted at model calibration, in the context of medical image segmentation.
Indeed, mL1-ACE will be much more sensitive to uncertain predictions such as those found on the periphery of segmented regions.

\begin{figure}[tbp]
	\centering
	\def\svgwidth{1\textwidth}
	\input{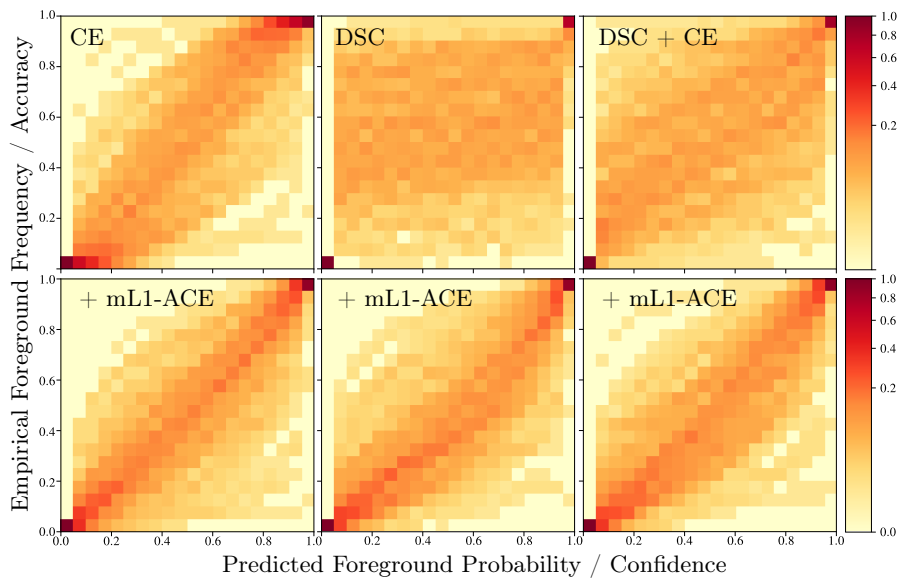}
	\caption{Dataset reliability histograms for BraTS 2021 dataset, with different loss functions, without (top row) and with (bottom row) mL1-ACE.}
	\label{fig:dataset_reliability_diagram}
\end{figure}

To capture reliability information at the dataset-level rather than image-level, we suggest to aggregate the image-specific metrics rather than computing a dataset-level DCA.
This not only provides an opportunity to gain insights in the variability within the dataset but also offers a pragmatic solution to deal with large datasets. 
To support visual assessment of dataset-level reliability,
we propose an extension of reliability diagrams illustrated in \figref{fig:dataset_reliability_diagram} which we refer to as dataset reliability histograms.
By binning the empirical foreground frequencies within each predicted foreground probability bin across all cases in a dataset, we create a joint histogram representation of reliability.
This provides an informative visualisation of calibration across an entire dataset.
\section{Experiments}
Training, validation and testing was performed using the Multimodal Brain Tumour Segmentation (BraTS) Benchmark training dataset from 2021 \cite{BraTS2021}.
Segmentations consist of three tumour sub-regions.
These sub-regions then form clinical tumour sub-regions, summarised as follows:
Enhancing tumour (ET) captures the gadolinium enhancement.
Tumour core (TC) is comprised of the necrotic tumour core (NCR) and ET.
Whole tumour (WT) is comprised of the TC and the peritumoral edematous and infiltrated tissue (ED).

The training dataset of $1251$ cases was randomly split into training, validation and testing with a ratio of $1000$:$51$:$200$ cases.
A basic UNet architecture \cite{ronneberger2015unet} implemented in MONAI \cite{MONAI} was used with $8$, $16$, $32$, and $64$ channels and strides of 2.
We use the same data augmentation pipeline, for training, as in~\cite{Fidon2022}, closely matching the one used for the nnUNet \cite{nnUNet}.

A training patch size of $224 \times 224 \times 144$ was used, with a batch size of $8$.
Training was ran for $1000$ epochs, with $16$ iterations per epoch.
The final model for evaluation was the one with the best validation DSC.
Different training strategies were compared, using combinations of popular segmentation losses with and without the mL1-ACE loss. We considered cross-entropy (CE), DSC, and CE + DSC loss functions.
Equal weighting for each loss function component was used.
For all calibration metrics and mL1-ACE loss, $20$ bins were used to discretise the predicted probability space.
For all runs, an Adam optimiser with a learning rate of $0.001$ was used.
Post-hoc Ts was performed for each model, using CE loss function, with the temperature parameter optimised using the validation dataset.
All experiments were performed using an NVidia DGX cluster with NVidia V100 (32GB) and Nvidia A100 (40GB) GPUs.

\section{Results}
We first consider the impact of our auxiliary loss in terms of segmentation performance and then focus on calibration.
The DSC scores for the different loss functions are summarised in \tabref{tab:mean_dice_summary}, showing that despite using a basic network with no hyper-parameter tuning, we obtain competitive DSC values on BraTS 2021.
The addition of mL1-ACE as an auxiliary loss with CE or DSC-CE losses, results in a small reduction in DSC scores (p < 0.01)  
However, for DSC loss, the addition of mL1-ACE maintains DSC performance, showing no statistically significant difference (p = 0.61).
The DSC values of models with post-hoc Ts are not included as Ts has no practically significant effect on DSC.

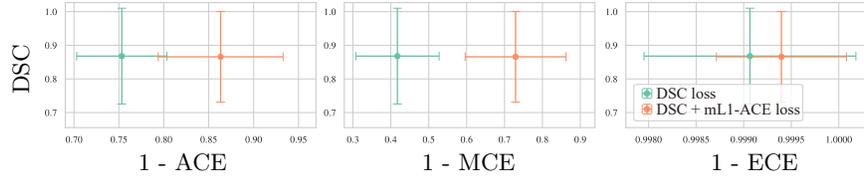
\begin{figure}[t]
	\centering
	\def\svgwidth{1\textwidth}
\begingroup%
  \makeatletter%
  \providecommand\color[2][]{%
    \errmessage{(Inkscape) Color is used for the text in Inkscape, but the package 'color.sty' is not loaded}%
    \renewcommand\color[2][]{}%
  }%
  \providecommand\transparent[1]{%
    \errmessage{(Inkscape) Transparency is used (non-zero) for the text in Inkscape, but the package 'transparent.sty' is not loaded}%
    \renewcommand\transparent[1]{}%
  }%
  \providecommand\rotatebox[2]{#2}%
  \newcommand*\fsize{\dimexpr\f@size pt\relax}%
  \newcommand*\lineheight[1]{\fontsize{\fsize}{#1\fsize}\selectfont}%
  \ifx\svgwidth\undefined%
    \setlength{\unitlength}{346.00000763bp}%
    \ifx\svgscale\undefined%
      \relax%
    \else%
      \setlength{\unitlength}{\unitlength * \real{\svgscale}}%
    \fi%
  \else%
    \setlength{\unitlength}{\svgwidth}%
  \fi%
  \global\let\svgwidth\undefined%
  \global\let\svgscale\undefined%
  \makeatother%
  \begin{picture}(1,0.22832368)%
    \lineheight{1}%
    \setlength\tabcolsep{0pt}%
    \put(0,0){\includegraphics[width=\unitlength,page=1]{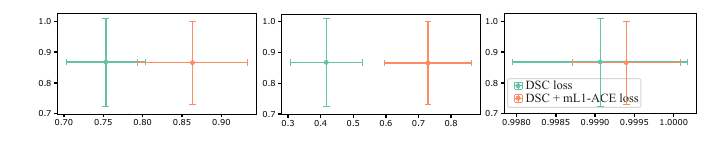}}%
    \put(0.21033797,0.025){\color[rgb]{0,0,0}\makebox(0,0)[t]{\lineheight{1.25}\smash{\begin{tabular}[t]{c}1 - ACE\end{tabular}}}}%
    \put(0.52197011,0.025){\color[rgb]{0,0,0}\makebox(0,0)[t]{\lineheight{1.25}\smash{\begin{tabular}[t]{c}1 - MCE\end{tabular}}}}%
    \put(0.83360225,0.025){\color[rgb]{0,0,0}\makebox(0,0)[t]{\lineheight{1.25}\smash{\begin{tabular}[t]{c}1 - ECE\end{tabular}}}}%
    \put(0.04302992,0.13574827){\color[rgb]{0,0,0}\rotatebox{90}{\makebox(0,0)[t]{\lineheight{1.25}\smash{\begin{tabular}[t]{c}DSC\end{tabular}}}}}%
  \end{picture}%
\endgroup%
	\caption{Comparative analysis of DSC against calibration metrics for average tumour component of BraTS 2021 dataset. Statistically significant difference (p < 0.01) for all metrics.}
	\label{fig:dice_versus_calibration}
\end{figure}

\begin{table}[tb]
	\caption{Dice Score across Loss Functions, higher is better}
	\label{tab:mean_dice_summary}
	\begin{tabularx}{\textwidth}{>{\hsize=1.2\hsize}X*{4}{>{\centering\arraybackslash\hsize=0.625\hsize}X}}
		\toprule
		Loss Function   & ET                   & TC                   & WT                   & Avg                  \\
		\midrule
		\rowcolor{gray!25}
		CE              & 0.82 ± 0.19          & 0.87 ± 0.18          & \textbf{0.90 ± 0.11} & 0.86 ± 0.14          \\
		CE + L1-ACE     & 0.80 ± 0.20          & 0.86 ± 0.19          & 0.87 ± 0.12          & 0.84 ± 0.15          \\
		\rowcolor{gray!25}
		DSC             & 0.84 ± 0.17          & 0.87 ± 0.20          & 0.89 ± 0.12          & 0.87 ± 0.14          \\
		DSC + L1-ACE    & 0.83 ± 0.18          & 0.88 ± 0.18          & 0.89 ± 0.11          & 0.87 ± 0.13          \\
		\rowcolor{gray!25}
		DSC-CE          & \textbf{0.85 ± 0.17} & \textbf{0.89 ± 0.17} & \textbf{0.90 ± 0.11} & \textbf{0.88 ± 0.13} \\
		DSC-CE + L1-ACE & 0.82 ± 0.20          & 0.86 ± 0.21          & 0.88 ± 0.13          & 0.85 ± 0.16          \\
		\bottomrule
	\end{tabularx}
\end{table}

We show the effect of Ts and mL1-ACE on the calibration of the models in \tabref{tab:average_calibration_error_summary}.
Our auxiliary loss has little effect on the calibration of the model trained with CE.
Our loss shows the biggest improvement in calibration when paired with the DSC loss function, with a 45\% reduction in ACE (p < 0.01). 
Overall the use of Ts shows marginal improvement in calibration, with the exception of CE loss where it has a small detrimental effect.
Similar conclusion can be drawn for MCE and ECE metrics, shown in Table S1 and Table S2 in the supplementary material.
We also note that the number of bins used to calculate our calibration metrics, has little effect, with ACE increasing by less than 3\% when increasing the bins from 10 to 100 (p < 0.01), as shown in supplementary Fig. S1.  

\begin{table}[t]
	\caption{Average Calibration Error across Loss Functions, lower is better}
	\label{tab:average_calibration_error_summary}
	\begin{tabularx}{\textwidth}{>{\hsize=1.2\hsize}X*{4}{>{\centering\arraybackslash\hsize=0.625\hsize}X}}
		\toprule
		Loss Function        & ET                   & TC                   & WT                   & Avg                  \\
		\midrule
		\rowcolor{gray!25}
		CE                   & 0.13 ± 0.09          & 0.14 ± 0.10          & 0.12 ± 0.07          & 0.13 ± 0.06          \\
		CE + Ts              & 0.15 ± 0.07          & \textbf{0.13 ± 0.08} & 0.13 ± 0.06          & 0.14 ± 0.05          \\
		CE + L1-ACE          & \textbf{0.12 ± 0.10} & 0.14 ± 0.10          & \textbf{0.11 ± 0.09} & \textbf{0.12 ± 0.07} \\
		CE + L1-ACE + Ts     & \textbf{0.12 ± 0.07} & \textbf{0.13 ± 0.08} & \textbf{0.11 ± 0.07} & \textbf{0.12 ± 0.05} \\
		\rowcolor{gray!25}
		DSC                  & 0.25 ± 0.07          & 0.26 ± 0.07          & 0.23 ± 0.06          & 0.25 ± 0.05          \\
		DSC + Ts             & 0.24 ± 0.07          & 0.24 ± 0.07          & 0.22 ± 0.06          & 0.24 ± 0.05          \\
		DSC + L1-ACE         & 0.14 ± 0.10          & 0.15 ± 0.09          & 0.12 ± 0.08          & 0.14 ± 0.07          \\
		DSC + L1-ACE + Ts    & \textbf{0.12 ± 0.09} & \textbf{0.13 ± 0.09} & \textbf{0.11 ± 0.07} & \textbf{0.12 ± 0.06} \\
		\rowcolor{gray!25}
		DSC-CE               & 0.20 ± 0.08          & 0.21 ± 0.08          & 0.18 ± 0.09          & 0.20 ± 0.06          \\
		DSC-CE + Ts          & 0.18 ± 0.09          & 0.18 ± 0.09          & 0.15 ± 0.09          & 0.17 ± 0.07          \\
		DSC-CE + L1-ACE      & 0.15 ± 0.11          & 0.15 ± 0.11          & \textbf{0.11 ± 0.08} & 0.14 ± 0.08          \\
		DSC-CE + L1-ACE + Ts & 0.13 ± 0.10          & 0.14 ± 0.10          & \textbf{0.11 ± 0.06} & 0.13 ± 0.07          \\
		\bottomrule
	\end{tabularx}
\end{table}

The relationship between DSC and calibration metrics is shown in \figref{fig:dice_versus_calibration}, for the DSC loss model, showing that the addition of mL1-ACE as an auxiliary loss maintains DSC performance with a substantial improvement in MCE and ACE (p < 0.01).  

\figref{fig:case_reliability_diagrams_dice} shows a reliability diagram for the DSC loss model, with and without mL1-ACE and Ts, for the WT segmentation component.
As can be seen, Ts has a limited effect on the calibration for an individual case, whereas the addition of mL1-ACE has a significant effect, clearly improving calibration.

\begin{figure}[tbp]
	\centering
	\includegraphics[width=1\textwidth]{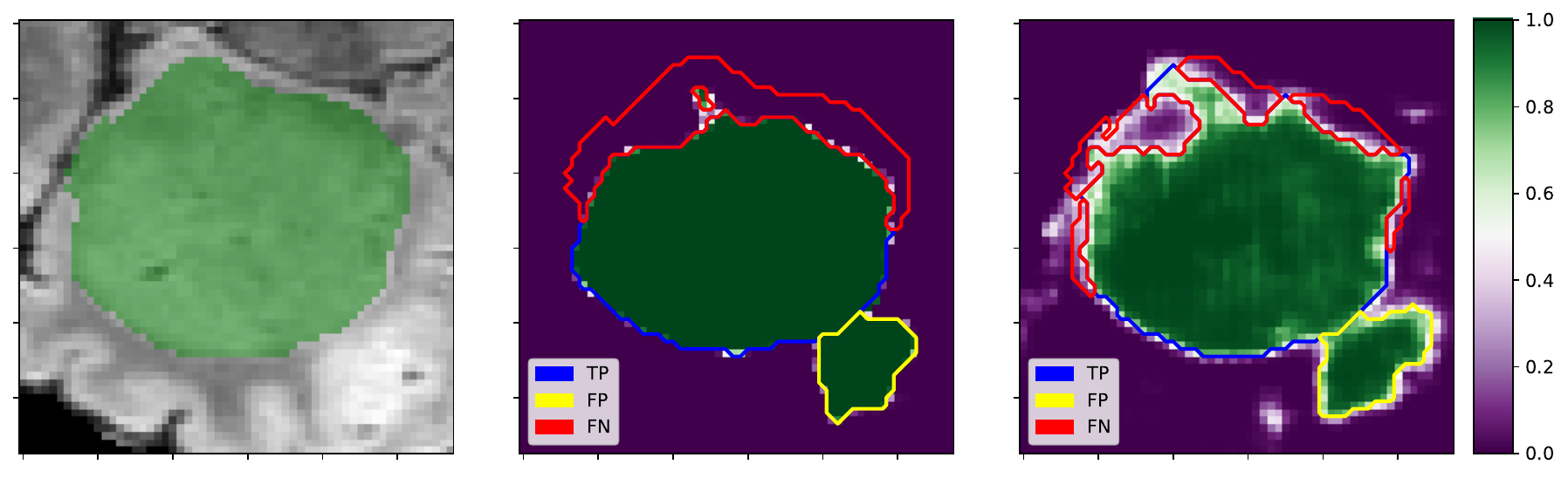}
	\caption{Segmentation of BraTS 2021 case 00014, with DSC loss function. Left: Ground truth WT on FLAIR image, Middle: Predicted WT foreground probabilities, Right: WT foreground probabilities with mL1-ACE. True positive, false positive, and false negative contours are shown in blue, yellow, and red, respectively.}
	\label{fig:segmentation_example}
\end{figure}

Similar, but more pronounced, results are shown in \figref{fig:dataset_reliability_diagram}, where the addition of mL1-ACE has a significant qualitative improvement on calibration of the model across the entire dataset, for all loss functions, showing a marked reduction in under- and over-confident cases, by the narrowing of the distribution along the diagonal. We supplement this visual qualitative analysis, by showing an individual case in \figref{fig:segmentation_example}, where the addition of mL1-ACE has a significant effect on the calibration of the model, with a marked reduction in the confidence of false positive and false negative predictions.

\section{Conclusion and discussion}
Prior work assumed that hard binning leads to non-differentiable DCA-based metrics.
We demonstrate that it is sufficiently differentiable and provide a multi-class and image-specific variant for use in challenging image segmentation tasks.
By performing averaging inside of each bin across a large enough number of voxels our approach indeed provides a sufficiently informative gradient updates. We further show that the introduction of the proposed mL1-ACE as an equally weighted auxiliary loss function to DSC loss, can improve the calibration of the model, with a $45$\% reduction in ACE and $55$\% reduction in MCE metrics, without any significant change in the Dice coefficient metric, remaining at competitive $87$\%, when evaluated on the BraTS 2021 dataset.

We also introduced the concept of dataset reliability histograms to visualise the reliability of a semantic segmentation models across an entire dataset, providing valuable insights into modes of failure.

Additionally, we show that models trained with cross-entropy loss tend to have much better calibration than those trained with a DSC-based loss, as has been reported in prior literature~\cite{yeung2021calibrating}.
Our method has limited benefit for CE-trained models. 
However, with DSC-based losses outperforming CE over a large set of tasks~\cite{lossodyssey2021}, the need for well calibrated DSC-based losses remains.
We also show that the post-hoc calibration method of temperature scaling has limited effect on the calibration of the model.

While our experiments were focused on BraTS with a baseline U-Net architecture, we believe that our mL1-ACE auxiliary loss has the potential to offer an easy-to-use and transferable solution to improve model calibration in any medical image segmentation approach. Exploring the use with other datasets, base losses, and architectures, including transformers, is left for future work.


\begin{credits}
\subsubsection{\ackname} The authors would like to acknowledge support for TB from EPSRC CDT [EP/S022104/1] and Intel, UK. 
TV is supported by a Medtronic / RAEng Research Chair [RCSRF1819$\backslash$7$\backslash$34]. 
This work was supported by core funding from the Wellcome/EPSRC [WT203148/Z/16/Z; NS/A000049/1]. 
For the purpose of open access, the authors have applied a CC BY public copyright licence to any Author Accepted Manuscript version arising from this submission. This article only uses publicly available datasets. Their re-use did not require any ethical approval.

\subsubsection{\discintname}
TV is co-founder and shareholder of Hypervision Surgical. 
BG is a part-time employee of Kheiron Medical Technologies and HeartFlow with stock options as part of the standard compensation package.
The authors have no other competing interests to declare that are relevant to the content of this article.
\end{credits}

\bibliographystyle{splncs04}
\bibliography{main}
\renewcommand{\thefigure}{S\arabic{figure}}
\renewcommand{\thetable}{S\arabic{table}} 
\setcounter{figure}{0}
\setcounter{table}{0}

\section*{Supplementary Material}

\begin{table}
	\caption{Maximum Calibration Error}
	\label{tab:mce_summary}
	\begin{tabularx}{\textwidth}{>{\hsize=1.2\hsize}X*{4}{>{\centering\arraybackslash\hsize=0.625\hsize}X}}
		\toprule
		                      & ET                   & TC                   & WT                   & Avg                  \\
		\midrule
		\rowcolor{gray!25}
		CE                    & 0.26 ± 0.19          & 0.28 ± 0.18          & 0.23 ± 0.12          & 0.25 ± 0.12          \\
		CE + Ts               & 0.30 ± 0.16          & 0.29 ± 0.16          & 0.25 ± 0.10          & 0.28 ± 0.10          \\
		CE + L1-ACE           & 0.24 ± 0.19          & 0.26 ± 0.18          & 0.20 ± 0.15          & {0.23 ± 0.14}        \\
		CE + L1-ACE + Ts      & \textbf{0.23 ± 0.15} & \textbf{0.25 ± 0.16} & 0.20 ± 0.12          & \textbf{0.23 ± 0.11} \\
		\rowcolor{gray!25}
		Dice                  & 0.59 ± 0.15          & 0.61 ± 0.15          & 0.55 ± 0.13          & 0.58 ± 0.11          \\
		Dice + Ts             & 0.56 ± 0.15          & 0.58 ± 0.15          & 0.51 ± 0.14          & 0.55 ± 0.11          \\
		Dice + L1-ACE         & 0.29 ± 0.18          & 0.30 ± 0.17          & 0.23 ± 0.14          & 0.27 ± 0.13          \\
		Dice + L1-ACE + Ts    & \textbf{0.23 ± 0.17} & \textbf{0.25 ± 0.16} & 0.21 ± 0.11          & \textbf{0.23 ± 0.12} \\
		\rowcolor{gray!25}
		Dice-CE               & 0.46 ± 0.17          & 0.46 ± 0.17          & 0.36 ± 0.17          & 0.43 ± 0.13          \\
		Dice-CE + Ts          & 0.26 ± 0.19          & 0.32 ± 0.19          & \textbf{0.19 ± 0.12} & 0.26 ± 0.13          \\
		Dice-CE + L1-ACE      & 0.29 ± 0.21          & 0.30 ± 0.20          & 0.21 ± 0.14          & 0.27 ± 0.15          \\
		Dice-CE + L1-ACE + Ts & 0.25 ± 0.19          & 0.27 ± 0.18          & 0.20 ± 0.11          & 0.24 ± 0.13          \\
		\bottomrule
	\end{tabularx}
\end{table}

\begin{table}
	\caption{Expected Calibration Error ($\times 10^{-4}$)}
	\label{tab:ece_summary}
	\begin{tabularx}{\textwidth}{>{\hsize=1.2\hsize}X*{4}{>{\centering\arraybackslash\hsize=0.625\hsize}X}}
		\toprule
		                      & ET                   & TC                   & WT                   & Avg                  \\
		\midrule
		\rowcolor{gray!25}
		CE                    & 4.21 ± 7.49          & 5.66 ± 10.5          & \textbf{9.71 ± 11.3} & 6.53 ± 7.53          \\
		CE + Ts               & 7.77 ± 7.16          & 10.2 ± 10.0          & 19.1 ± 9.45          & 12.4 ± 6.93          \\
		CE + L1-ACE           & 4.00 ± 4.75          & 6.16 ± 9.41          & 12.0 ± 12.9          & 7.38 ± 6.94          \\
		CE + L1-ACE + Ts      & 21.1 ± 4.69          & 27.2 ± 6.65          & 62.0 ± 11.1          & 36.8 ± 5.69          \\
		\rowcolor{gray!25}
		Dice                  & 5.09 ± 7.67          & 7.48 ± 16.5          & 15.5 ± 17.8          & 9.34 ± 11.1          \\
		Dice + Ts             & 4.95 ± 7.63          & 7.36 ± 16.5          & 15.0 ± 17.8          & 9.10 ± 11.1          \\
		Dice + L1-ACE         & 3.47 ± 7.56          & 4.50 ± 8.45          & 10.1 ± 12.8          & \textbf{6.03 ± 6.84} \\
		Dice + L1-ACE + Ts    & \textbf{3.27 ± 7.17} & \textbf{4.41 ± 8.00} & 12.0 ± 10.9          & 6.54 ± 5.84          \\
		\rowcolor{gray!25}
		Dice-CE               & 4.40 ± 7.57          & 5.70 ± 12.2          & 11.6 ± 14.4          & 7.21 ± 8.62          \\
		Dice-CE + Ts          & 3.84 ± 7.43          & 5.20 ± 11.8          & 10.2 ± 13.4          & 6.41 ± 8.11          \\
		Dice-CE + L1-ACE      & 3.91 ± 8.09          & 6.04 ± 14.6          & 10.3 ± 14.8          & 6.75 ± 9.84          \\
		Dice-CE + L1-ACE + Ts & 4.23 ± 7.68          & 6.56 ± 14.0          & 13.9 ± 12.6          & 8.23 ± 8.83          \\
		\bottomrule
	\end{tabularx}
\end{table}

\begin{figure}[tb]
	\centering
	\def\svgwidth{1\textwidth}
\begingroup%
  \makeatletter%
  \providecommand\color[2][]{%
    \errmessage{(Inkscape) Color is used for the text in Inkscape, but the package 'color.sty' is not loaded}%
    \renewcommand\color[2][]{}%
  }%
  \providecommand\transparent[1]{%
    \errmessage{(Inkscape) Transparency is used (non-zero) for the text in Inkscape, but the package 'transparent.sty' is not loaded}%
    \renewcommand\transparent[1]{}%
  }%
  \providecommand\rotatebox[2]{#2}%
  \newcommand*\fsize{\dimexpr\f@size pt\relax}%
  \newcommand*\lineheight[1]{\fontsize{\fsize}{#1\fsize}\selectfont}%
  \ifx\svgwidth\undefined%
    \setlength{\unitlength}{345.82676625bp}%
    \ifx\svgscale\undefined%
      \relax%
    \else%
      \setlength{\unitlength}{\unitlength * \real{\svgscale}}%
    \fi%
  \else%
    \setlength{\unitlength}{\svgwidth}%
  \fi%
  \global\let\svgwidth\undefined%
  \global\let\svgscale\undefined%
  \makeatother%
  \begin{picture}(1,0.28688525)%
    \lineheight{1}%
    \setlength\tabcolsep{0pt}%
    \put(0,0){\includegraphics[width=\unitlength,page=1]{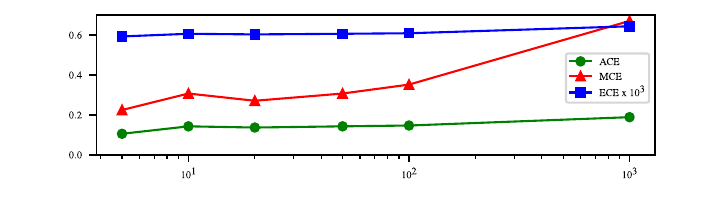}}%
    \put(0.50274294,0.0141168){\color[rgb]{0.14117647,0.10980392,0.10980392}\makebox(0,0)[t]{\lineheight{1.25}\smash{\begin{tabular}[t]{c}Number of Bins\end{tabular}}}}%
    \put(0.07523192,0.16295043){\color[rgb]{0.14117647,0.10980392,0.10980392}\rotatebox{90}{\makebox(0,0)[t]{\lineheight{1.25}\smash{\begin{tabular}[t]{c}Metric Value\end{tabular}}}}}%
  \end{picture}%
\endgroup%
	\caption{Comparison of calibration metrics using different number of bins when evaluated on the DSC + mL1-ACE loss model. Both ACE and ECE are very stable with respect to number of bins (5, 10, 20, 50, 100, 1000). MCE shows a larger increase from 100 to 1000 bins, this is due to the presence of empty bins, a likely occurrence when using such a high number of bins.\label{fig:bin_no}}
\end{figure}


\end{document}